\documentclass{bmvc2k}

\usepackage{multirow}
\usepackage{array}

\title{Rethinking Classification and Localization for Cascade R-CNN}

\addauthor{Ang Li}{angless@njust.edu.cn}{1}
\addauthor{Xue Yang}{yangxue-2019-sjtu@sjtu.edu.cn}{2}
\addauthor{Chongyang Zhang}{zcy603@163.com}{1}

\addinstitution{
 Nanjing University of Science and Technology \\
 Nanjing, China \\
}
\addinstitution{
 Shanghai Jiao Tong University \\
 Shanghai, China\\
}

\runninghead{Li ET AL.}{Rethinking Classification and Localization for Cascade R-CNN}

\def\etal{\emph{et al}\bmvaOneDot}

\begin{document}
\maketitle

\begin{abstract}

We extend the state-of-the-art Cascade R-CNN with a simple feature sharing mechanism. Our approach focuses on the performance {\em increases on high IoU but decreases on low IoU thresholds}---a key problem this detector suffers from. Feature sharing is extremely helpful, our results show that given this mechanism embedded into all stages, we can easily narrow the gap between the last stage and preceding stages on low IoU thresholds {\em without resorting to the commonly used testing ensemble} but the network itself. We also observe obvious improvements on all IoU thresholds benefited from feature sharing, and the resulting cascade structure can easily  match or exceed its counterparts, only with negligible extra parameters introduced. To push the envelope, we demonstrate 43.2 AP on COCO object detection without any bells and whistles including testing ensemble, surpassing previous Cascade R-CNN by a large margin. Our framework is easy to implement and we hope it can serve as a general and strong baseline for future research.

\end{abstract}

\section{Introduction}
\label{sec:intro}

\begin{figure*}[htbp]
\centering
\subfigure{
\includegraphics[width=8cm]{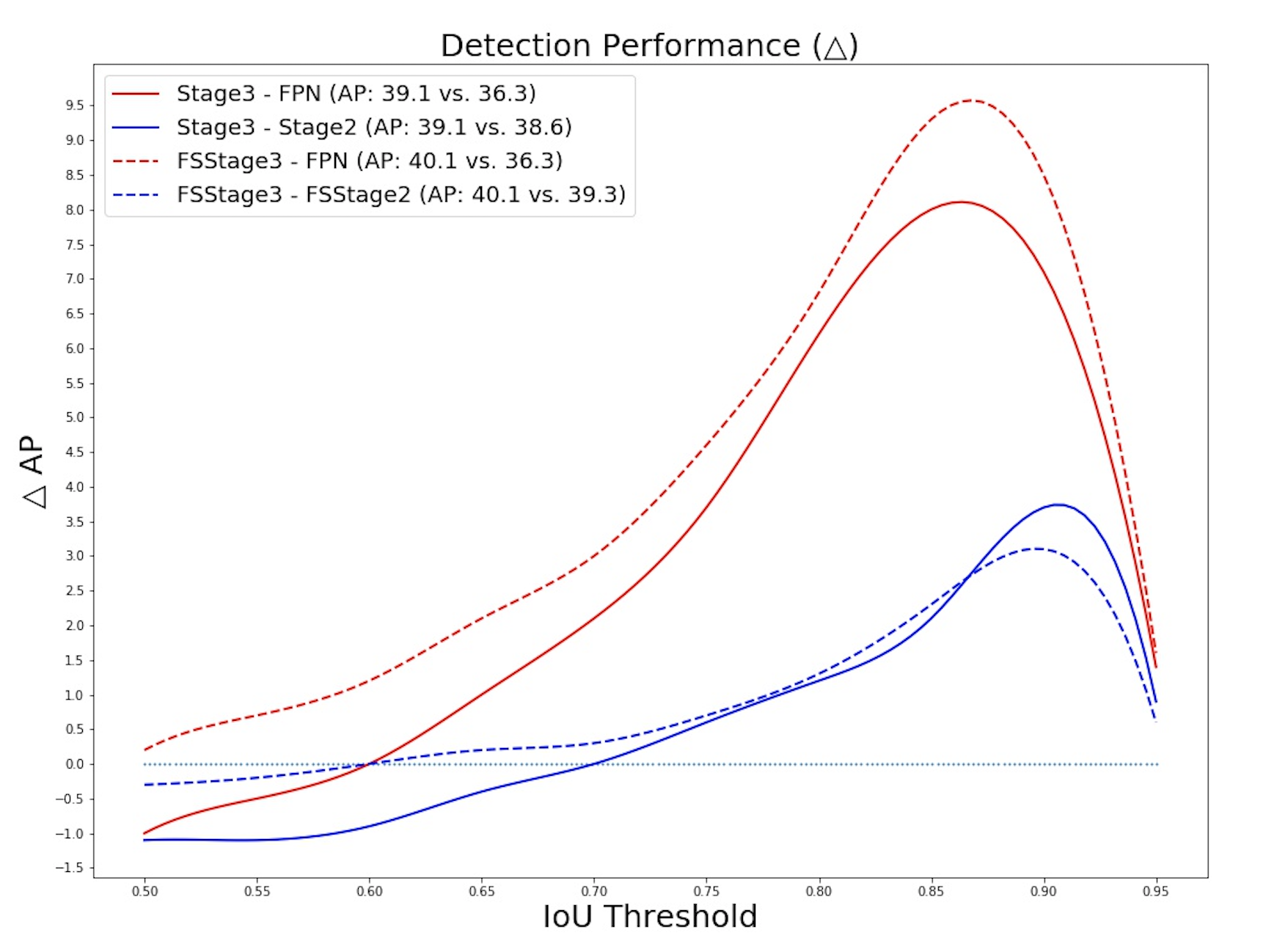}
}
\caption{\small{Solid line: the difference of Stage3 minus FPN, and Stage3 minus stage2 on the AP value ($\triangle$ AP) between IoU 0.5 and 0.95 for original Cascade RCNN. Dashed line: the differene of FSStage3 minus FPN, and FSStage3 minus FSStage2 on the AP value ($\triangle$ AP) for our FSCascade. We show the results based on ResNet-50 and `1$\times$' training strategy. Our approach is more close to zero on low IoU thresholds ({\em lower gap}) and much more improvement on high IoU thresholds ({\em higher peak}).}}
\label{fig:image1}
\end{figure*}

Object detection is a fundamental yet challenging task in computer vision, requiring the algorithm to predict a bounding box with a category label for each object of interest in an image. Current state-of-the-art approaches\cite{cai2018cascade,lin2017feature,ren2015faster} resort to a two-stage, anchor-refined manner, which applies the pre-defined anchor boxes to generate the region proposals first, and then refines the sampling RoIs progressively. Cascade R-CNN\cite{cai2018cascade} follows these thoughts and proposes to refine the RoIs with gradually rising foreground/background thresholds to make the network more concentrate on high IoU settings. However, as can be clearly seen from Figure~\ref{fig:image1}, for the 3rd stage, Cascade R-CNN achieves suboptimal performance on low IoU thresholds compared with the 2nd stage and FPN\cite{lin2017feature}, and this gap is even larger for the former. In original Cascade R-CNN\cite{cai2018cascade}, the best results are achieved by ensembling of three classifiers on the 3rd stage proposals. 

This gap may be obvious because, for the 3rd stage, a higher IoU threshold 0.7 is usually employed to distinguish between the foreground and background in the training phase, thus the network will give low confidence with high probability to the RoIs whose IoU between 0.5 and 0.7 and will not regress these RoIs. This bias means some low IoU RoIs may learn well in preceding stages but may not in the 3rd stage, resulting in too much {\em high overlap but low confidence} boxes, and finally aborted by the NMS procedure. The commonly used testing ensemble process can alleviate this because in the testing phase, when the network can not acknowledge the overlaps (with ground truth) in advance, averaging the scores of all stages can make up for this problem of low confidence in the 3rd stage,  meaning that the network can still recall the aforementioned boxes with a certain probability, while mataining the quantities of original high IoU boxes, leading to more balanced results on all IoU thresholds.

{\em Can we achieve identical results without testing ensemble?} Or in other words, can we narrow the gap between the last stage and preceding stages on low IoU threholds while keeping the superior performance on high IoU thresholds? We empirically find the testing ensemble process could be encoded in the network itself as long as effectively making use of the feature sharing mechanism among all stages. Feature sharing in Cascade R-CNN\cite{cai2018cascade} has the following advantages, firstly, given this embedded into classification, this score averaging process can be inserted into the network, which can be viewed as a `soft' ensemble approach, thus the current stage can also learn from preceding stages without depending on each stage's isolated prediction, making the scoring  become more reasonable. Secondly, given this embedded into localization, the current stage can refine the RoIs progressively with the help of preceding stages, and the jointly optimization through back-propagation can make the detection box more accurate, thus the overall performance becomes improved.

Based on the analysis above, we propose FSCascade, a simple feature sharing Cascade R-CNN framework, aiming at narrowing the gap between the last stage and preceding stages on low IoU thresholds without resorting to the testing ensemble, while improving the overall performance benefited from feature sharing. Based on FSCascade, we can clearly see from Figure~\ref{fig:image1} that our approach not only brings overall improvements but also outperforms FPN\cite{lin2017feature} baseline on low IoU thresholds, and more importantly, compared with its own 2nd stage and 3rd stage, FSCascade greatly reach this goal. Moreover, extensive experiments also demonstrate employing testing ensemble approach to pray for a final best performance is very tricky, requiring carefully adjusting the parameters. However, our method is more stable and can improve the final performance steadily. On COCO object detection task, FSCascade surpasses all previous single-model detectors by a large margin, achieving 43.2 AP without any bells and whistles including testing ensemble, the main contributions of this work are summarized as follows:

\begin{itemize}
\item[-] We give an in-depth analysis of why Cascade R-CNN is not ideal on low IoU thresholds.
\item[-] We propose a soft ensemble approach called FSCascade, improving the generic object detection performance on all IoU thresholds without depending on the tricky post-processing but the network itself, and the improvements are much more stable.
\item[-] We achieve 43.2 AP on COCO object detection task using ResNet-101 without any bells and whistles including testing ensemble, surpassing previous stage-of-the-art approach Cascade R-CNN by a large margin.
\end{itemize}

\begin{figure*}[htbp]
\centering
\subfigure[CFS]{
\includegraphics[width=5.5cm]{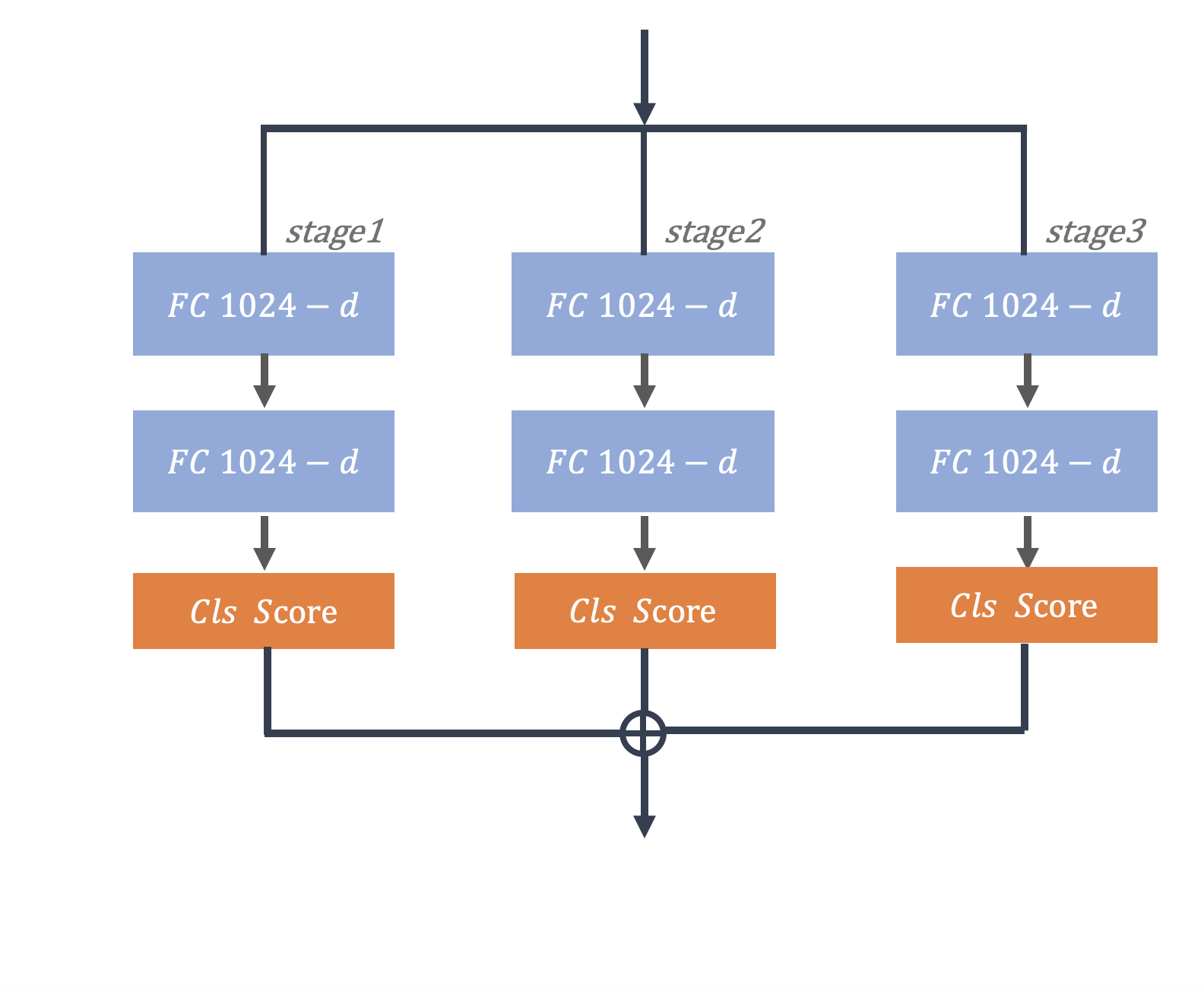}
}
\quad
\subfigure[LFS]{
\includegraphics[width=5.5cm]{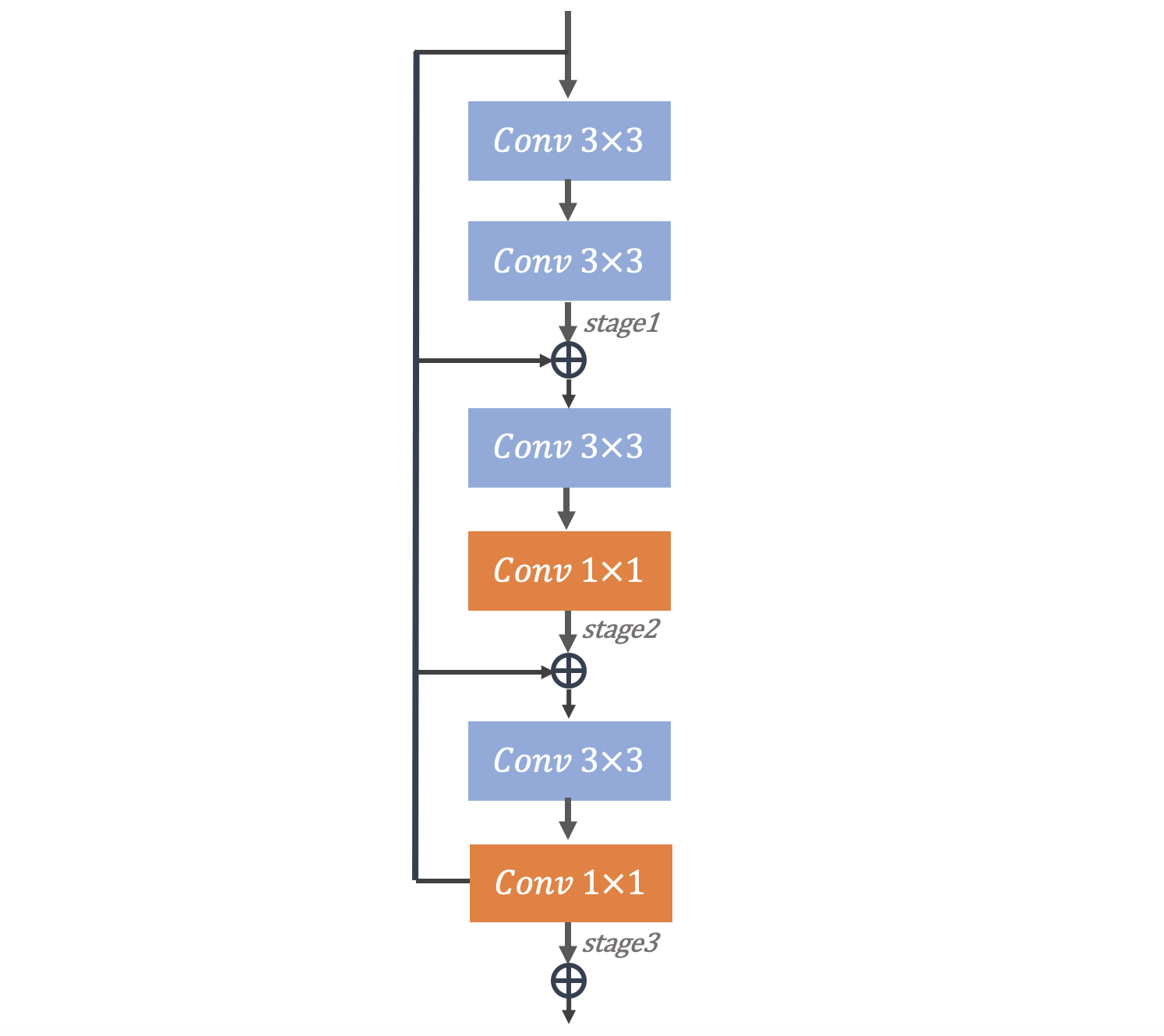}
}

\caption{\small{Structures of classification feature sharing (CFS) and localization feature sharing (LFS) shown in the left and right, respectively. The CFS is executed in a parallel way and the LFS in a serial manner. We show the sharing structure of the 3rd stage for simplicity.}}
\label{fig:image2}
\end{figure*}

\section{Related Works}
\label{sec:related}

Multi-stage object detectors are very popular in recent years. Following the main idea of ``divide and conquer', these detectors optimze a simpler problem first, and then refine the difficult progressively. In the field of object detection, cascade can be introduced into two components, namely the proposal generation process usually called `RPN' and the classification and localization predicting process usually called `R-CNN\cite{girshick2014rich}'. In these works, for the former, \cite{yang2016craft,gidaris2016attend,zhong2017cascade} propose to use a multi-stage procedure to generate accurate proposals, and then refine these proposals with a single Fast R-CNN\cite{girshick2015fast}. For the latter, Cascade R-CNN\cite{cai2018cascade} is the most famous object detectors among them, with increasingly foreground/background thresholds selected to refine the ROIs progressively. HTC\cite{chen2019hybrid} follows this thoughts and propose to refine the features in an interleaved manner, resulting in state-of-the-art performance on instance segmentation task. Other works like \cite{gidaris2015object,li2015convolutional} also apply the R-CNN stage several times, but the performance is far behind Cascade R-CNN\cite{cai2018cascade}.

Feature sharing has also been taken in many approaches. In \cite{lin2017feature}, sharing features in RPN stage can improve the performance and similar results can be found in \cite{lin2017focal,chen2019hybrid} across different tasks. Different from these methods, our approach not only focuses on the {\em overall improvements} but also {\em narrowing the gap} without resorting to the commonly used testing ensemble in cascaded approaches but the network itself based on feature sharing.

\section{Methodology}
\label{sec:method}

Our goal is to improve the last stage performance both on low and high IoU thresholds with minimal modifications, so special network design is not our purpose. We follow the main structure of Cascade R-CNN based on FPN and reach this goal with the main idea of feature sharing both for classification and localization, with only negligible extra parameters introduced, discussed next.

\subsection{Classification}

Object detection needs to assign a category label for each object instance. In Cascade R-CNN, each stage is responsible for predicting a confidence score for a specific class under a given foreground/background threshold. What we find in this process is given a low IoU box (between 0.5 and 0.7), it may obtain an extremely high score in preceding stages, but a lower score in the following stage. For the last stage, if the probability of scoring the low IoU box can be increased, then in the testing phase, more these boxes could be recalled. That is to say, for the low IoU boxes, the key problem of the last stage is that {\em a given box may already regress well but may not obtain a reasonable confidence score}, and we need to give this ability back.

Performing classification feature sharing is straightforward because, in FPN, two 1024-d fully connected layers are more friendly to this task. So for the classification in each stage, we directly employ these two layers as sharing base, and the sharing structure is shown in the left of Figure~\ref{fig:image2}. For a given stage, the pooled features (7$\times$7 in our experiments) are not only passed through its own fully connected layers, but also passed through all preceding two fully connected layers trained on lower IoU, the entire process is executed in parallel, and the simple implementation is shown below:
\begin{equation}
\begin{aligned}
  C_1  &= W_1^2 (W_1^1 (X_i^{cls}))\\  
  C_2  &= W_2^2 (W_2^1 (X_i^{cls}))\\
  \vdots\\  
  C_i  &= W_i^2 (W_i^1 (X_i^{cls})).\\
\end{aligned} 
\end{equation}
Here, $W_i^1$ and $W_i^2$ denote the feature transformation of two fully connected layers for stage $i$, $X_i^{cls}$ denotes the pooled features, and $C_i$ is added with the features trained on all preceding stages through element-wise sum to represent final classification features, then the predictions are performed.

This feature sharing mechanism can be viewed as a soft ensemble method, however, different from the traditional testing ensemble, our approach can benefit from smoother information flow across different stages and optimize for all IoU thresholds through back-propagation, making the training process more stable.

\begin{figure*}[htbp]
\centering
\subfigure[Cascade R-CNN]{
\includegraphics[width=4.8cm]{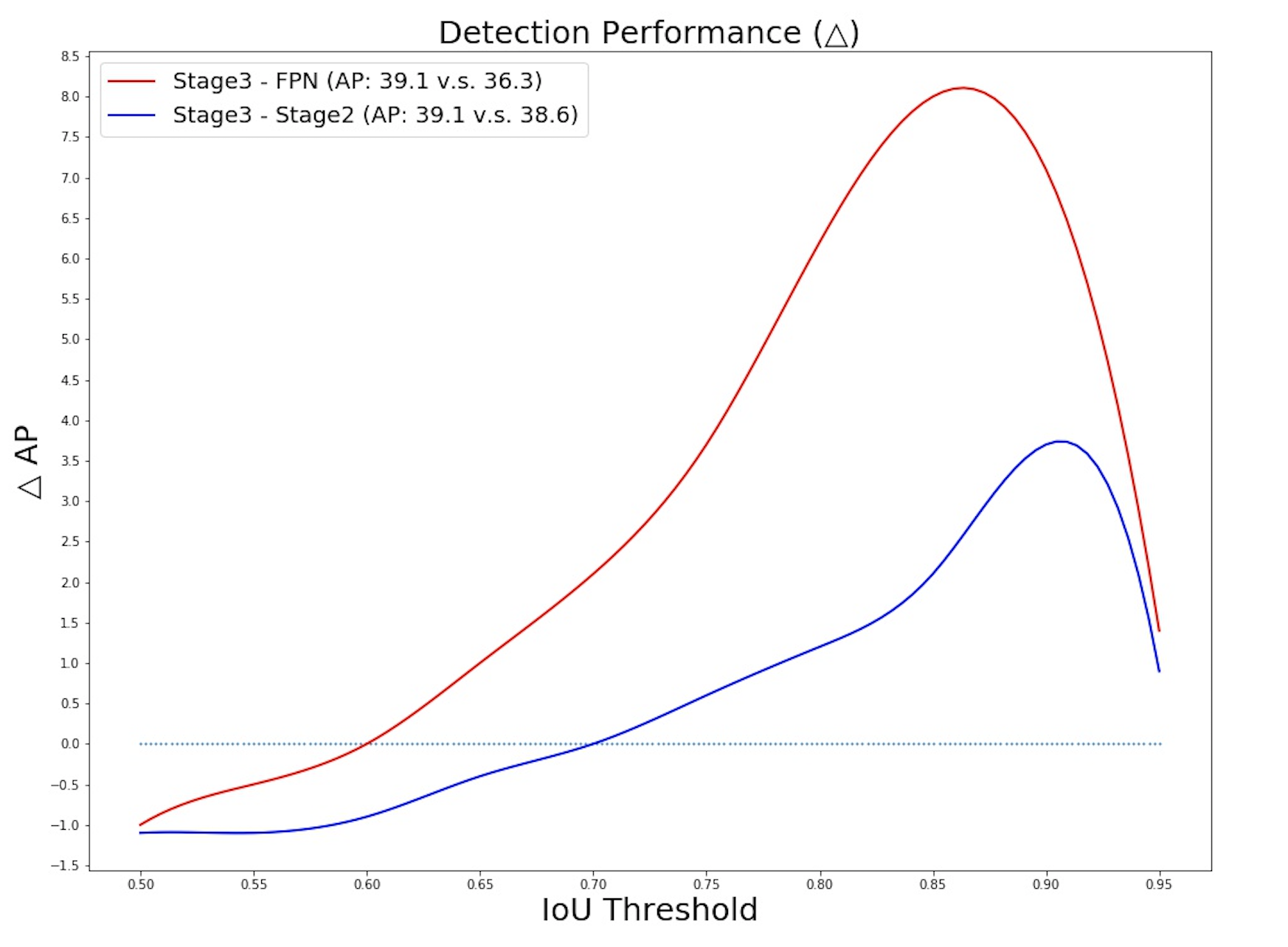}
}
\quad
\subfigure[CFS]{
\includegraphics[width=4.8cm]{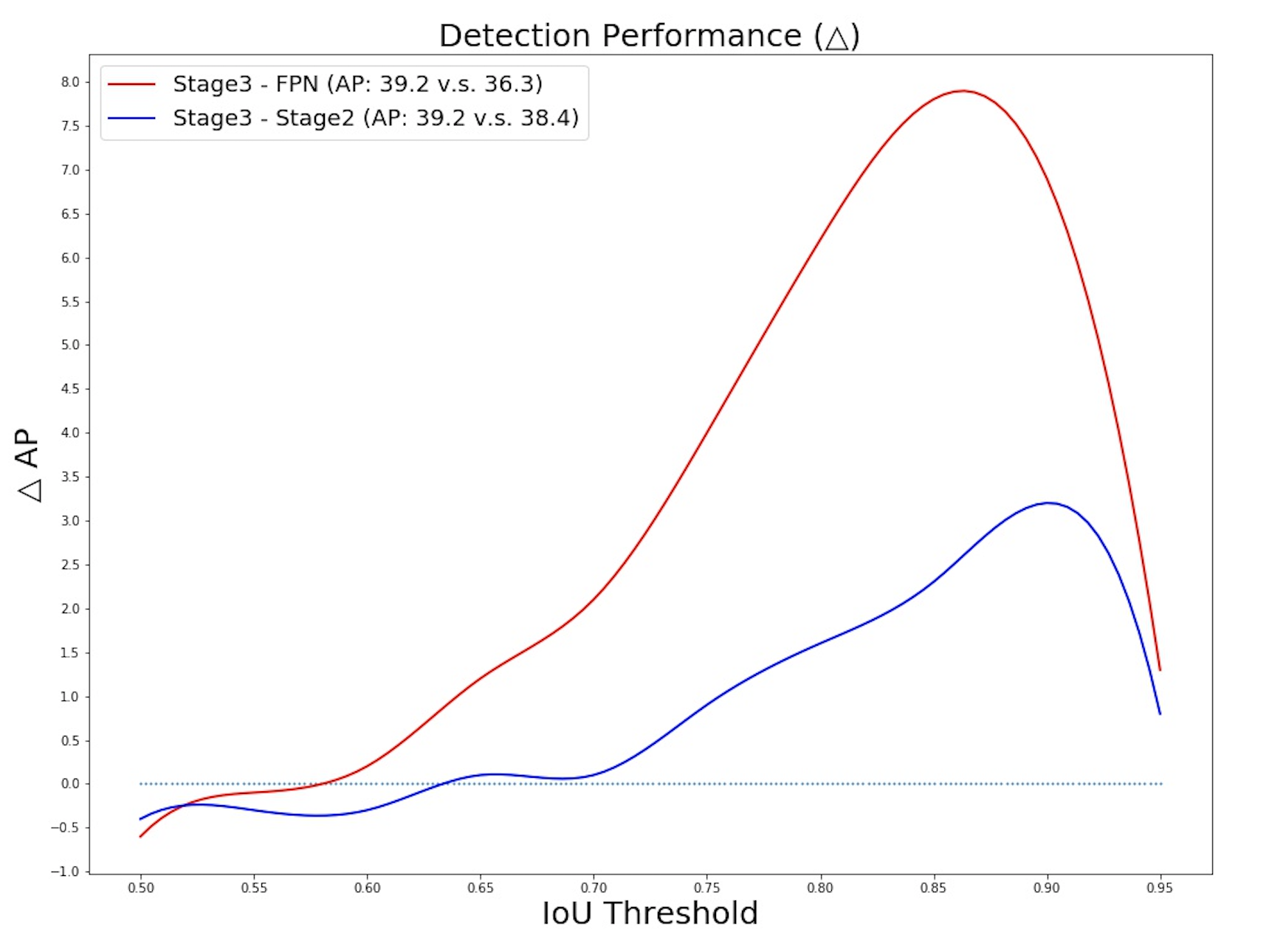}
}
\quad
\subfigure[LFS]{
\includegraphics[width=4.8cm]{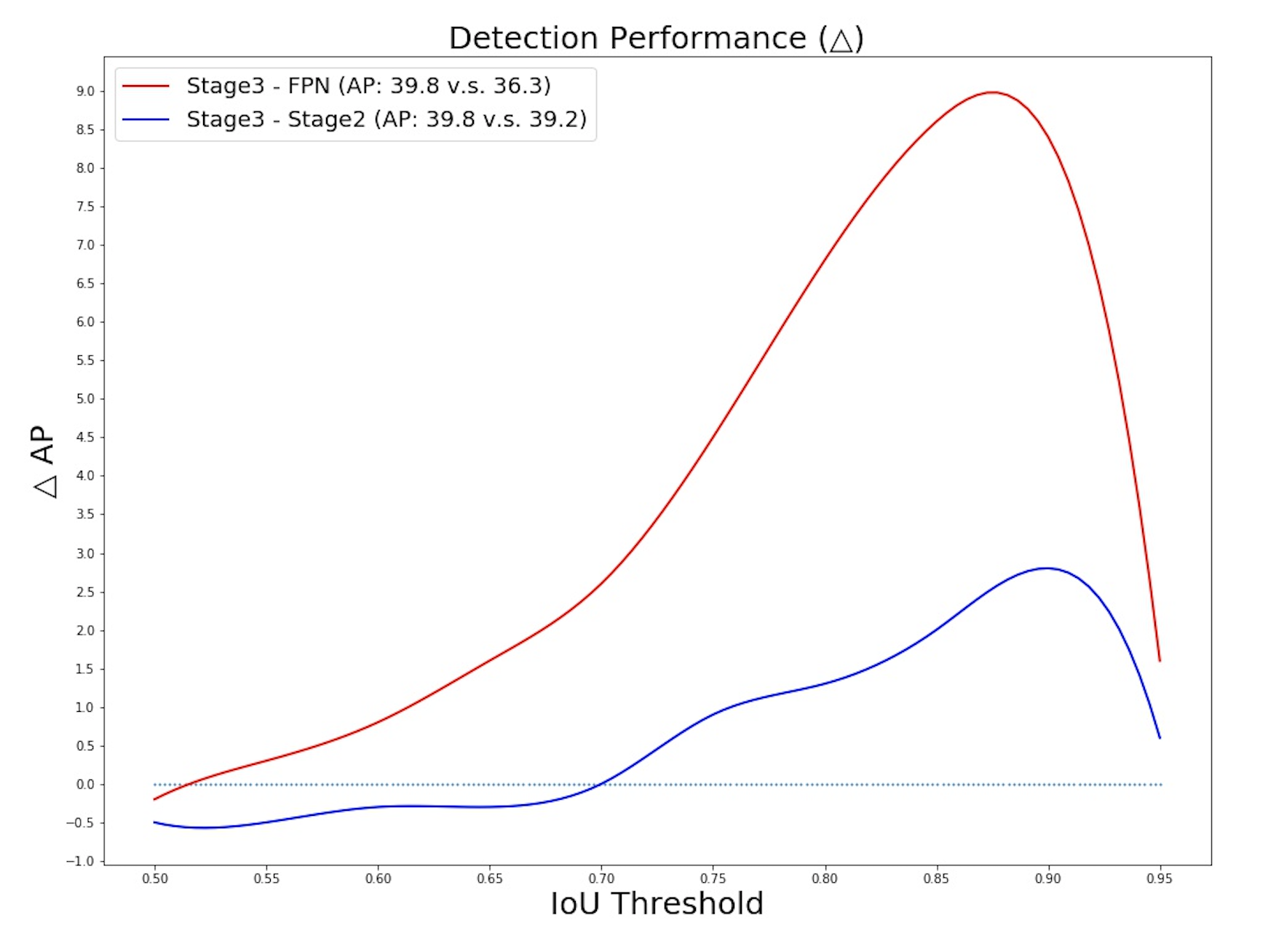}
}
\quad
\subfigure[FSCascade]{
\includegraphics[width=4.8cm]{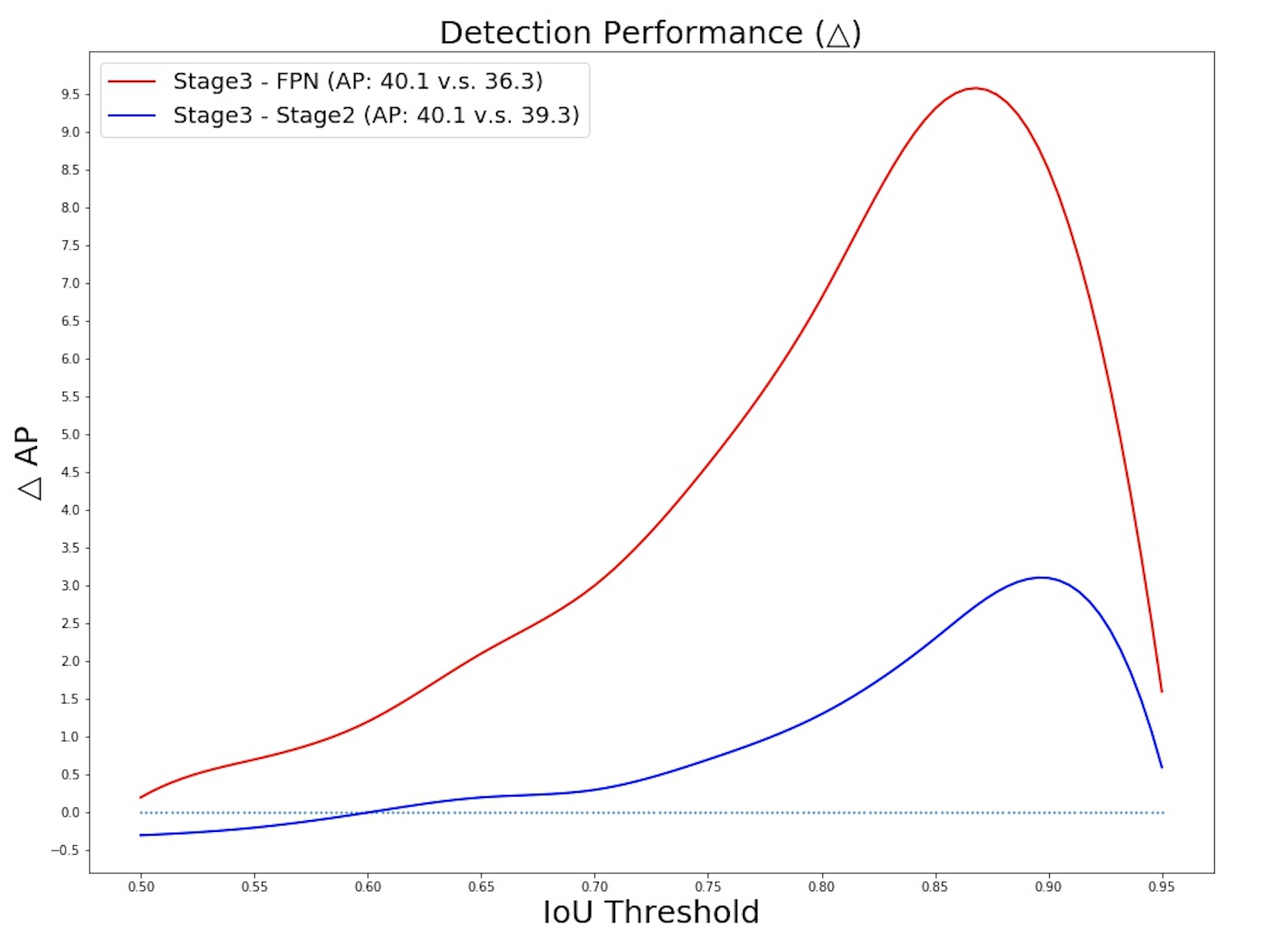} 
}
\caption{\small{The difference of stage3 minus FPN, and stage3 minus stage2 on the AP value ($\triangle$ AP) for all the experiments based on ResNet-50 with `1$\times$' training strategy. (a) is original Cascade R-CNN, (b) is FSCascade only with classification features shared (CFS). (c) is FSCascade only with localization features shared (LFS), and (d) is our FSCascade. }}
\label{fig:image3}
\end{figure*}

\subsection{Localization}

Solved the problem of imbalanced scoring for the last stage, our next consideration is that will the overall performance can be further improved while keeping this trend of the narrowed gap? To reach this goal, we firstly propose to perform localization based on the {\em convolutional head} instead of the commonly used fully connected layers used in FPN, because localization needs fine spatial information preserved. Due to this decoupled prediction head, we empirically find it can improve the detection performance by a large margin, shown in the experiments section next.

However,  it is worth noting although the added convolutional layers are more friendly to localization, it still makes the aforementioned problems---largely gap between the last stage and preceding stages on low IoU thresholds, unsolved. That is to say, the added layers may improve the localization performance but {\em may not narrow the gap}. However, with the help of feature sharing, we can reach this goal.

Different from classification features sharing in a parallel way, we perform feature sharing for localization in a serial manner based on the added convolutional layers, shown in the right of Figure~\ref{fig:image2}. The reason behind this is that box features are more friendly to be refined progressively based on preceding stages. As can be seen from Figure~\ref{fig:image2}, two 3$\times$3 convolutional layers are added to the 1st stage, and for the following stage, only one 3$\times$3 and one 1$\times$1 convolutional layer are added to each stage. All the added layers are 256 channels to be compatible with original structures. Our principle is trying to make the convolutional head as simple as possible without increasing too much computational overhead. The implementation is shown below:

\begin{equation}
\begin{aligned}
  B_1  &= F_1^2 (F_1^1 (X_i^{box}))\\  
  B_2  &= X_i^{box} + G_1(F_2^1 (B_1))\\
  \vdots\\
  B_i  &= X_i^{box} + G_{i-1}(F_i^1 (B_{i-1})).\\
\end{aligned} 
\end{equation}

Here, $F_i$ denotes feature transformations of the $3 \times 3$ convolutional layer, $G_i$ denotes feature transformations of the $1 \times 1$ convolutional layer, and $B_i$ is used to perform box predictions. We also experiment with the parallel way like classification, but the results are no better than the serial manner, which further verifies that feature sharing for classification and localization are fundamentally different, the former more concentrates on `balance' while the latter more emphasizes `accuracy'.

\begin{figure*}[htbp]
\label{fig:fig1}
\centering
\subfigure{
\includegraphics[width=6cm,height=4.8cm]{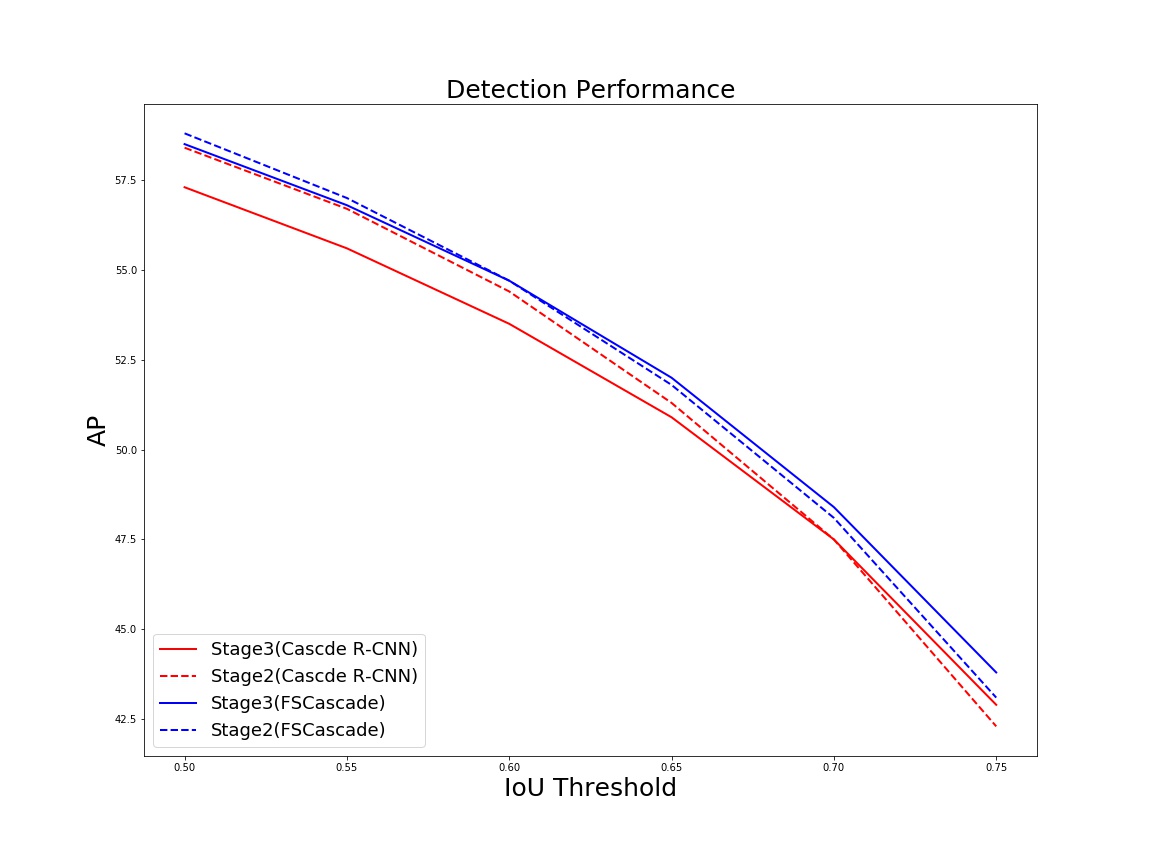} 
}
\caption{\small{The overall performance on COCO {\tt val} sets of our FSCascade {\em  vs.} Cascade R-CNN based on ResNet-50 with `1$\times$' training strategy. The IoU thresholds between {\bf 0.5} and {\bf 0.75} are shown above to verify: (1) the narrowed gap (blue dashed line {\em vs.} blue solid line), (2) the overall improvements (red solid line {\em vs.} blue solid line).}}
\label{img:image4}
\end{figure*}

\section{Experiments}
\label{sec:exp}

We perform experiments on MS-COCO 2017 datasets, all models were trained on the $\sim$118k training sets, evaluated on the 5k validation sets ({\tt val}), and final results were also reported on the $\sim$20k testing sets ({\tt test-dev}). We use standard COCO-style Average Precision (AP) metric with IoU settings ranging from 0.5 to 0.95 with an interval of 0.05, and box AP is used to measure the detection performance. We use Cascade R-CNN based on FPN backbone as our baseline.

\subsection{Implementation Details}
\label{sec:sec41}

For a fair comparison, we reimplemented {\em all the experiments including baseline} with the same codebase, unless specifically stated. Our baseline results are slightly higher than the reported performance in the original papers. For Cascade R-CNN, we adopt a 3-stage cascade, and follow all the settings with the official Detectron version except the training strategy, discussed next. We train all the detectors with 8 GPUs (two images per GPU for both ResNet-50 and ResNet-101) for 20 epochs with the initial learning rate of 0.02, we decay it by 0.1 for the first epoch to perform warm up, and reset it back, then decay it by 0.1 after 10 and 16 epochs. The longer edge and shorter edge of images are resized to 1200 and 800, the same with the testing size. All ablation studies are based on ResNet-50.

\begin{figure*}[htbp]
\centering
\subfigure[Cascade R-CNN]{
\includegraphics[width=6cm]{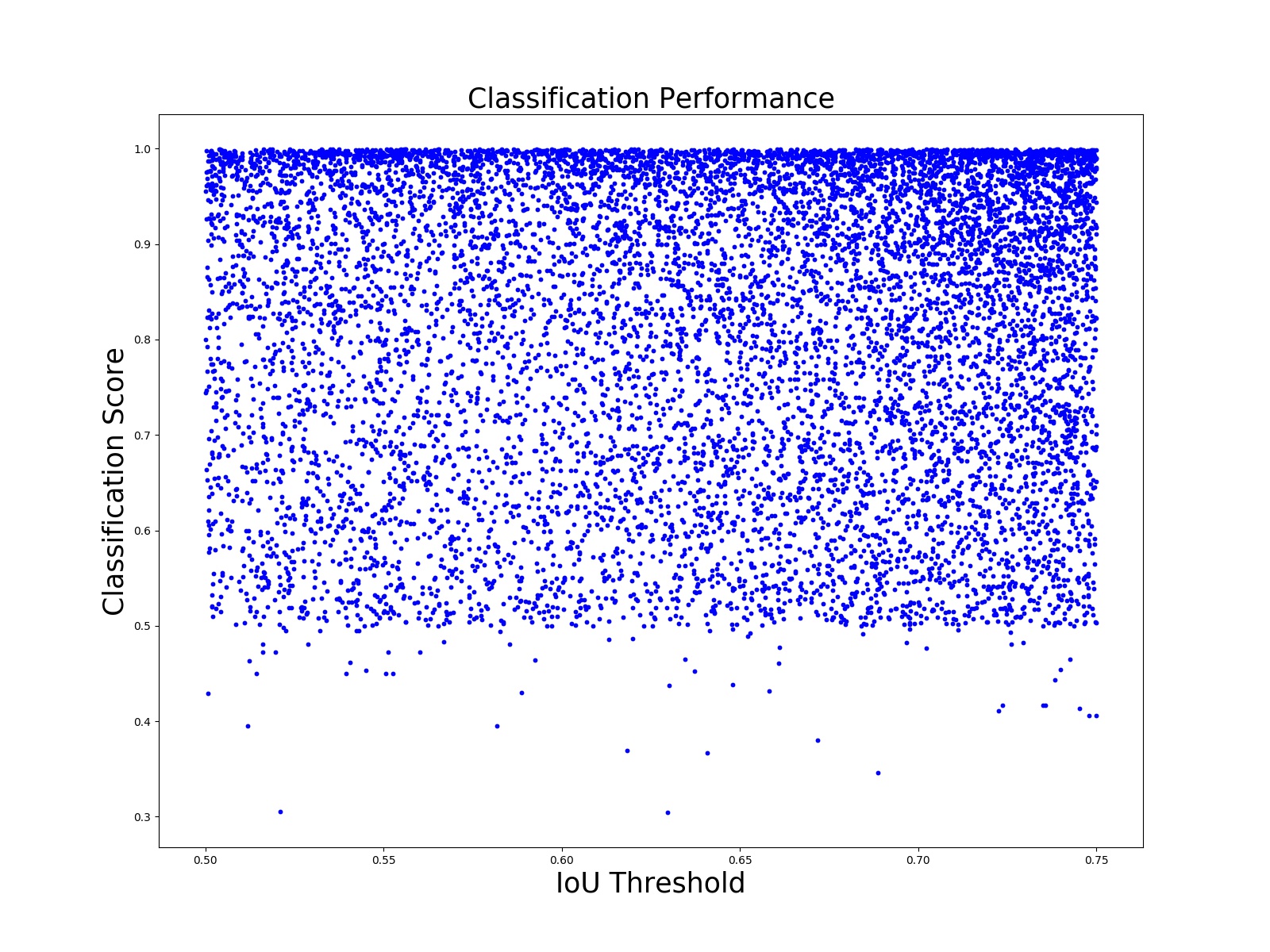}
}
\quad
\subfigure[FSCascade]{
\includegraphics[width=6cm]{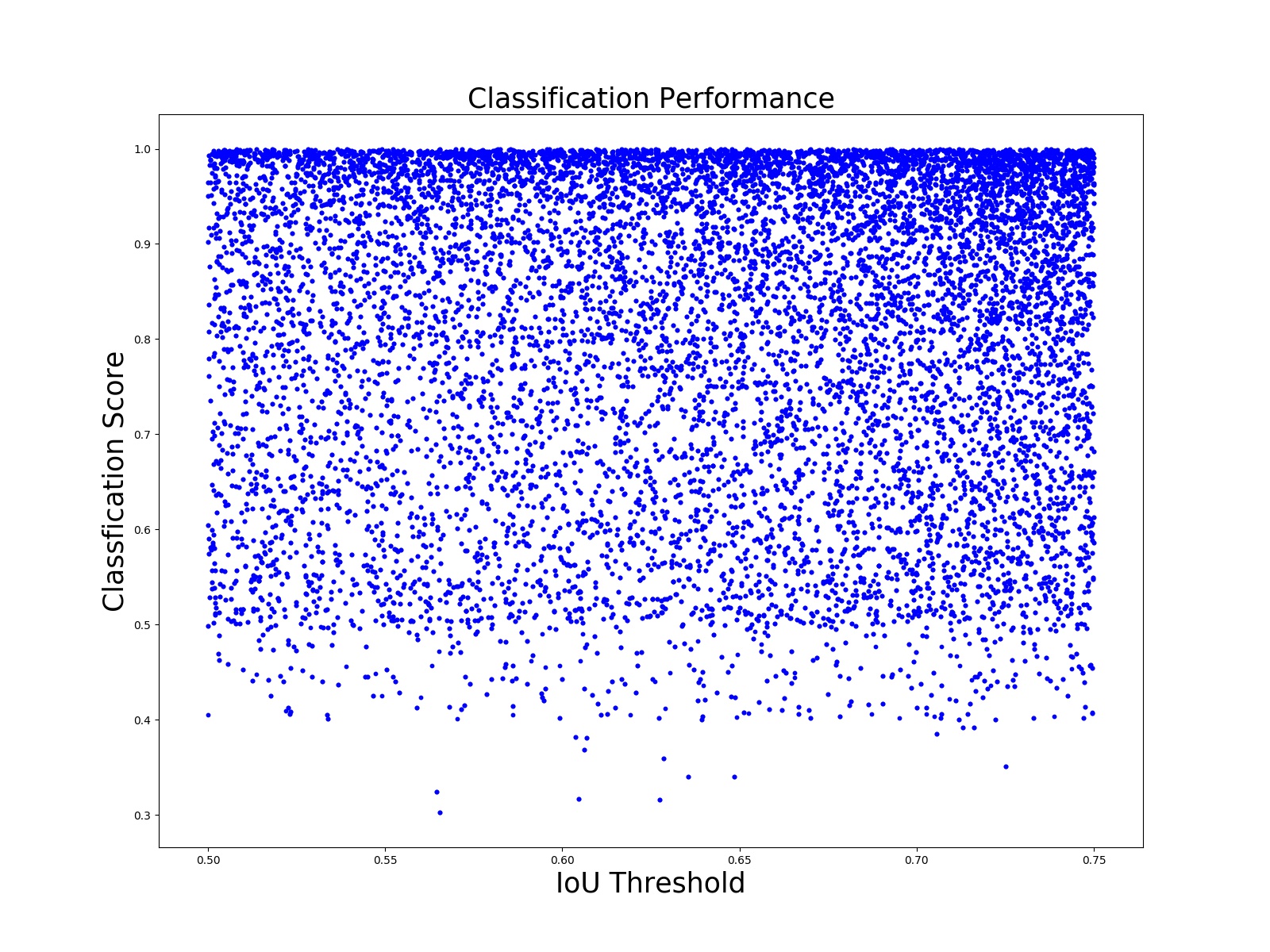}
}
\caption{\small{Confidence scores of the detection boxes whose IoU with ground truth boxes between {\bf 0.5} and {\bf 0.75} on the COCO {\tt val} sets, more balanced scores can be seen in the right image (b) based on FSCascade. }}
\label{img:image5}
\end{figure*}

\subsection{Ablation Studies}

{\bf Classification feature sharing ({\em CFS}) is essential.} In Figure~\ref{fig:image3} we verify the effectiveness of feature sharing for classification. Based on (a) and (b), we can clearly see that models with {\em vs.} without feature sharing have similar APs, and even a little better for the latter (39.1 {\em vs.} 39.2). However, their gap for the 3rd stage with FPN (-1.0 {\em vs.} -0.6 for AP$_{50}$), and the 3rd stage with the 2nd stage (-1.1 {\em vs.} -0.4 for AP$_{50}$), showing a big difference. Based on the experiments, we come to the following conclusions. Firstly, CFS makes the scoring process of the last stage more reasonable, especially for low IoU boxes, thus the gap is extremely narrowed. Secondly, CFS may not contribute too much to the overall performance, which can be complementary to LFS to some extent.

We also emphasize that these improvements introduce no extra parameters, with only `soft' ensemble taken in parallel.

{\bf Localization feature sharing ({\em LFS}) is essential.} We also verify the effectiveness of feature sharing for localization in Figure~\ref{fig:image3}, and similar conclusions observed. Firstly, benefiting from the convolutional layers on preserving spatial information, the overall performance has been greatly improved (39.1 {\em vs.} 39.8). Secondly, comparing (a) with (c), we can clearly see LFS further narrows the gap for the 3rd stage with FPN (-1.0 vs. -0.2 for AP$_{50}$), and more importantly, the 3rd stage with the 2nd stage (-1.1 {\em vs.} -0.5 for AP$_{50}$).

\begin{table*}
\footnotesize
\renewcommand{\arraystretch}{1.12}
\begin{center}
\begin{tabular}{p{2cm}<{\centering}|p{0.56cm}p{0.56cm}p{0.56cm}p{0.56cm}p{0.56cm}p{0.56cm}p{0.56cm}p{0.56cm}p{0.56cm}p{0.56cm}}

Method & AP$_{50}$ & AP$_{55}$ & AP$_{60}$ & AP$_{65}$ & AP$_{70}$ & AP$_{75}$ & AP$_{80}$ & AP$_{85}$ & AP$_{90}$ & AP$_{95}$\\
\hline
Cascade R-CNN & 57.3 & 55.6	& 53.5 & 50.9 & 47.5 & 42.9 & 36.9 & 28.5  & 15.7 & 2.2\\
\hline
Stage3 (w/o LFS) & 58.0 & 56.2 & 54.1 & 51.4 & 48.1 & 43.7 & 37.9 & 29.4 & 17.0 & 2.3 \\ 
Stage2 (w/o LFS) & 58.8 & 56.9 & 54.9 & 51.8 & 47.8 & 42.5 & 35.8 & 27.3 & 13.8 & 1.4 \\ 
$\triangle$ & {\em -0.8} & {\em -0.7} & {\em -0.8} & {\em -0.4} & {\em -0.3} & {\em +1.2} & {\em +2.1}  & {\em +2.1}  & {\em +3.2}  & {\em +0.9}  \\
\hline
Stage3 (w/ LFS) & 58.2 & 56.4 & 54.2 & 51.5 & 48.0 & 43.7 & 37.5 & 29.1 & 17.0 & 2.4 \\ 
Stage2 (w/ LFS) & 58.7 & 56.9 & 54.6 & 51.8 & 48.0 & 42.8 & 36.2 & 27.1 & 14.1 & 1.8 \\ 
$\triangle$ & {\em -0.5} & {\em -0.5} & {\em -0.4} & {\em -0.3} & {\em 0.0} & {\em +0.9} & {\em +1.3}  & {\em +2.0}  & {\em +2.9}  & {\em +0.6}  \\
\end{tabular}
\end{center}
\caption{\small{Training with {\em vs.} without localization feature sharing both based on convolutional localization head, evaluated on COCO {\tt val} sets. The CFS is not taken in this experiment. Cascade R-CNN indicates the 3rd stage results without testing ensemble.}}
\label{tab:table1}
\end{table*}

{\em Is this narrowed gap only caused by the added convolutional layers?} For the answer, we also experiment to verify with {\em vs.} without feature sharing based on the {\em convolutional localization head}, and the results are shown in Table~\ref{tab:table1}. This table reveals that convolutional head is more friendly to localization (39.1 {\em vs.} 39.8), but does not necessarily the {\em key} factor to encode this `soft' ensemble into the network itself (-0.8 {\em vs.} -0.5 for AP$_{50}$). 

{\bf The importance of feature sharing.} With the dynamic combination of CFS and LFS, FSCascade significantly narrows the gap between the last stage and preceding stages on low IoU thresholds while improving the overall performance, respectively. We visualize the overall performance (Average Precision) between IoU {\bf 0.5} and {\bf 0.75} based on FSCascade and our reimplemented Cascade R-CNN to verify the aforementioned conclusions and the results are shown in Figure~\ref{img:image4}.

Finally, for the last stage, we visualize the confidence scores of each detection box whose IoU with ground truth boxes between {\bf 0.5} and {\bf 0.75}, and the results are shown in Figure~\ref{img:image5}. We can clearly see that for a given IoU threshold, FSCascade makes the scoring process more reasonable, and the overall confidence scores are more balanced, thus many low IoU boxes can still be recalled, resulting in the final improvements on low IoU thresholds without resorting to the testing ensemble.

\begin{table*}
\footnotesize
\renewcommand{\arraystretch}{1.12}
\begin{center}
\begin{tabular}{p{2.9cm}<{\centering}|p{3cm}<{\centering}|p{0.56cm}p{0.56cm}p{0.56cm}<{\centering}|p{0.56cm}p{0.56cm}p{0.56cm}}

Method & Backbone & AP & AP$_{50}$ & AP$_{75}$ & AP$_{S}$  & AP$_{M}$ & AP$_{L}$ \\
\hline
{\em One-Stage Detectors:} \\
\hline
RetinaNet\cite{lin2017focal} & ResNet-101 & 39.1 & 59.1 & 42.3 & 21.8 & 42.7 & 50.2 \\
GHM\cite{li2018gradient} & ResNet-101 & 39.9 & 60.8 & 42.5 & 20.3 & 43.6 & 54.1 \\
CornerNet\cite{law2018cornernet} & Hourglass-104 & 40.5 & 56.5 & 43.1 & 19.4 & 42.7 & 53.9 \\
FSAF\cite{zhu2019feature} & ResNet-101 & 40.9 & 61.5 & 44.0 & 24.0 & 44.2 & 51.3 \\
\hline
{\em Two-Stage Detectors:} \\
\hline
Deformable R-FCN\cite{dai2017deformable} & Aligned-Inception-ResNet & 37.5 & 58.0 & 40.8 & 19.4 & 40.1 & 52.5 \\
Mask R-CNN\cite{he2017mask} & ResNet-101 & 38.2 & 60.3 & 41.7 & 20.1 & 41.1 & 50.2 \\
Soft-NMS\cite{bodla2017soft} & Aligned-Inception-ResNet & 40.9 & 62.8 & - & 23.3 & 43.6 & 53.3 \\
Grid R-CNN w/FPN\cite{DBLP:journals/corr/abs-1811-12030} & ResNeXt-101 & 43.2 & 63.0 & 46.6 & 25.2 & 46.5 & 55.2 \\
\hline
{\em Multi-Stage Detectors}: \\
\hline
AttractioNet\cite{gidaris2016attend} & VGG16+Wide ResNet & 35.7 & 53.4 & 39.3 & 15.6 & 38.0 & 52.7 \\
Cascade R-CNN\cite{cai2018cascade}$^{*}$& ResNet-50 & 40.6 & 59.9 & 44.0 & 22.6 & 42.7 & 52.1 \\
Cascade R-CNN\cite{cai2018cascade}$^{*}$& ResNet-101 & 42.8 & 62.1 & 46.3 & 23.7 & 45.5 & 55.2 \\
\hline
Cascade R-CNN$^{\dag}$(ours) & ResNet-50 & 40.2 & 58.8 & 44.0 & 22.3 & 42.5 & 51.9 \\
Cascade R-CNN$^{\dag}$(ours) & ResNet-101 & 42.3 & 61.4 & 46.6 & 24.5 & 45.3 & 53.6\\
FSCascade$^{\dag}$(ours) & ResNet-50 & 41.7 & 60.5 & 45.9 & 24.3 & 44.3 & 53.1 \\
FSCascade$^{\dag}$(ours) & ResNet-101 & {\bf 43.2} & 62.3 & 47.5 & 25.1 & 46.2 & 54.8\\

\end{tabular}
\end{center}
\caption{\small{Comparison with the state-of-the-art single-model detectors on COCO {\tt test-dev}. ``*'' denotes the ensemble of three stages. ``$\dag$'' denotes only the last stage used without testing ensemble.}}
\label{tab:table2}
\end{table*}

\subsection{Comparison with the state-of-the-art}

We use FSCascade based on FPN and ResNet-101 as backbone, to compare with state-of-the-art {\em single model} object detectors, and the results are shown in Table~\ref{tab:table2}. We follow all the training settings described in Section~\ref{sec:sec41}, only except extending the training time which usually called `2$\times$' training strategy. We do not use any multi-scale training or testing strategy. Three groups of detectors on Table~\ref{tab:table2} are one-stage, two-stage, and multi-stage, respectively. Our FSCascade with ResNet-101 backbone achieves 43.2 AP without any bells and whistles including testing ensemble (contributing to more than 0.5AP improvements as described in original Cascade R-CNN\cite{cai2018cascade} paper with this technique). We can also clearly see that with the solely 3rd stage, our FSCascade can match the performance of Grid R-CNN with FPN and ResNeXt-101\cite{xie2017aggregated} backbone and exceed the testing ensemble Cascade R-CNN by 2.7 \% (original implementation) and 3.5 \% (our reimplementation) for ResNet-50, 0.9 \% (original implementation) and 1.6\% (our reimplementation) for ResNet-101, respectively. 

\section{Conclusion}
\label{sec:conc}

In this paper, we propose FSCascade, a simple extension to the standard Cascade R-CNN with feature sharing mechanism. With our FSCascade, we can: (1) narrow the gap between the last stage and all preceding stages on low IoU thresholds, while (2) improving the overall 
performance on all IoU thresholds, with only negligible extra parameters introduced. We hope our simple approach can serve as a general and strong baseline for future research.

\section*{Acknowledgement}
This work is partially supported by National Key R\&D Program of China (No.2017YFB1300205).

\bibliography{egbib}
\end{document}